\documentclass[sigconf,natbib=true]{acmart}

\usepackage{graphicx}
\usepackage{multirow}
\usepackage{subcaption}
\usepackage{xcolor}
\usepackage{pifont}

\AtBeginDocument{%
  }

\setcopyright{acmlicensed}
\copyrightyear{2025}
\acmYear{2025}
\setcopyright{acmlicensed}\acmConference[SIGIR '25]{Proceedings of the 48th International ACM SIGIR Conference on Research and Development in Information Retrieval}{July 13--18, 2025}{Padua, Italy}
\acmBooktitle{Proceedings of the 48th International ACM SIGIR Conference on Research and Development in Information Retrieval (SIGIR '25), July 13--18, 2025, Padua, Italy}
\acmDOI{10.1145/3726302.3730291}
\acmISBN{979-8-4007-1592-1/2025/07}

\def\tcr{\textcolor{red}}
\def\tcb{\textcolor{blue}}
\begin{document}

\title{\includegraphics[height=0.6cm]{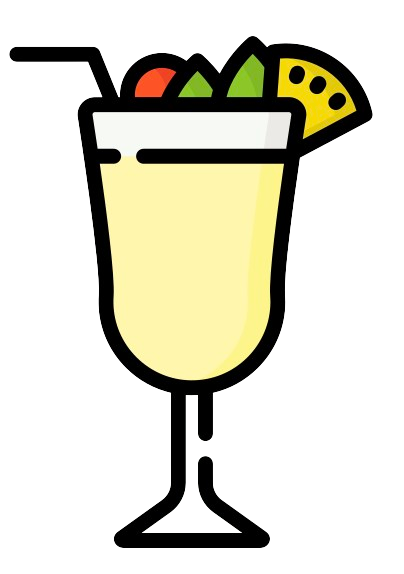} 
     CoLoTa: A Dataset for Entity-based Commonsense Reasoning over Long-Tail Knowledge}

\author{Armin Toroghi}
\affiliation{%
  \institution{University of Toronto}
  \city{Toronto}
  \state{ON}
  \country{Canada}
}
\email{armin.toroghi@mail.utoronto.ca}

\author{Willis Guo}
\affiliation{%
  \institution{Carnegie Mellon University}
  \city{Pittsburgh}
  \state{PA}
  \country{USA}}
  \email{willisg@andrew.cmu.edu}

\author{Scott Sanner}
\affiliation{%
  \institution{University of Toronto}
  \city{Toronto}
  \state{ON}
  \country{Canada}
}
 \email{ssanner@mie.utoronto.ca}


\renewcommand{\shortauthors}{Toroghi et al.}

\begin{abstract}
The rise of Large Language Models (LLMs) has redefined the AI landscape, particularly due to their ability to encode factual and commonsense knowledge, and their outstanding performance in tasks requiring reasoning. Despite these advances, hallucinations and reasoning errors remain a significant barrier to their deployment in high-stakes settings. In this work, we observe that even the most prominent LLMs, such as OpenAI-o1, suffer from high rates of reasoning errors and hallucinations on tasks requiring commonsense reasoning over obscure, long-tail entities. To investigate this limitation, we present a new dataset for \underline{Co}mmonsense reasoning over \underline{Lo}ng-\underline{Ta}il entities (CoLoTa), that consists of 3,300 queries from question answering and claim verification tasks and covers a diverse range of commonsense reasoning skills. We remark that CoLoTa can also serve as a Knowledge Graph Question Answering (KGQA) dataset since the support of knowledge required to answer its queries is present in the Wikidata knowledge graph. However, as opposed to existing KGQA benchmarks that merely focus on factoid questions, our CoLoTa queries also require commonsense reasoning. Our experiments with strong LLM-based KGQA methodologies indicate their severe inability to answer queries involving commonsense reasoning. Hence, we propose CoLoTa as a novel benchmark for assessing both (i) LLM commonsense reasoning capabilities and their robustness to hallucinations on long-tail entities and (ii) the commonsense reasoning capabilities of KGQA methods.
\end{abstract}





\maketitle

\section{Introduction}
The emergence of Large Language Models (LLMs) has marked a transformative phase
in AI deployment. This is largely due to their vast factual and commonsense knowledge, and their ability to reason by leveraging this knowledge. A key advantage offered by these models is their capability of entity-based commonsense reasoning, i.e., commonsense reasoning about specific objects, places, people, etc., that has made them suited for interacting with the real world. However, the issue of \textit{hallucination}—generating content that contradicts ground truth facts—along with reasoning errors has hindered the practical applicability of LLMs, especially in high-stakes applications. Despite the recent progress in enhancing the reasoning capabilities of LLMs, we observe that even advanced models such as Open AI-o1, GPT-4o, Gemini-1.5 flash, and Llama-3.3 
show high rates of reasoning errors and hallucinations in performing commonsense reasoning to answer queries about unfamiliar entities. 

Existing LLMs have shown strong performance on existing entity-based commonsense reasoning benchmarks such as StrategyQA~\citep{strategyqa} and CREAK~\citep{onoe2021creak} that focus on popular entities such as \textit{``Barack Obama''} and \textit{``François Mitterrand''}. Abundant information about these entities exists online and is included in the training data of LLMs. Hence, LLMs are well-equipped with the factual knowledge required to answer questions like \textit{``Could Barack Obama and François Mitterrand have met while they both held the position of president?"}. Using their commonsense reasoning skills, many LLMs can easily identify the necessity of comparing the time these two people held the position of \textit{``president''} to provide the correct answer. However, when the same commonsense question is asked about two less well-known entities such as \textit{``Liau Hiok-hian''} and \textit{``Virginia Raggi''}, instead of refraining from answering or querying for factual information, even state-of-the-art LLMs begin to generate hallucinated facts and commit commonsense reasoning errors.

\begin{figure*}[!t]
  \centering
  \includegraphics[width=0.85\linewidth]{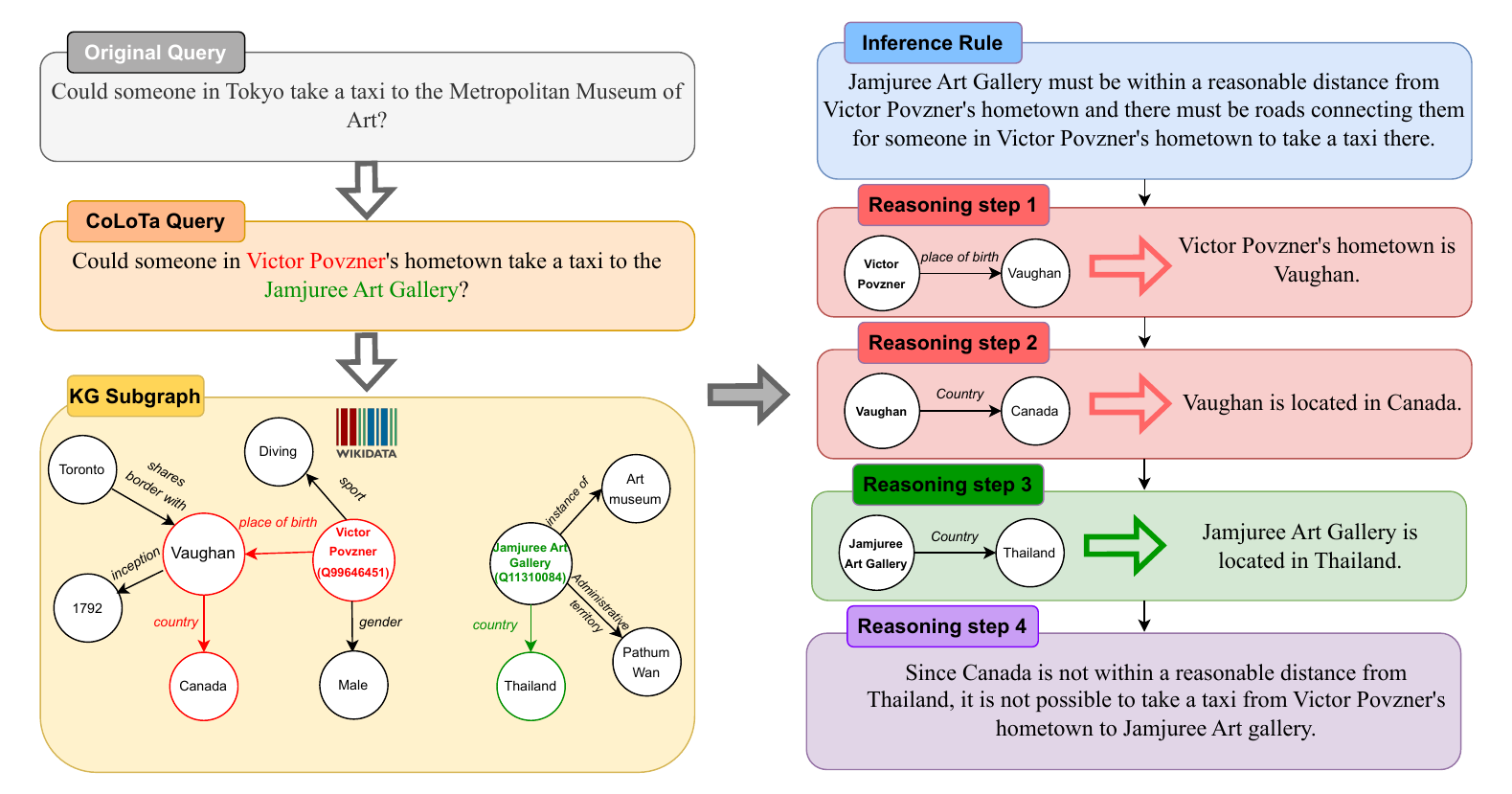}
    \caption{Constituents of an entry from CoLoTa: (i) an entity-based commonsense reasoning query (ii) Wikidata QIDs of the anchor entities, (iii) relevant Wikidata sub-graph that contains factual information to answer the query, (iv) an inference rule establishing the commonsense reasoning required to answer the query, and (v) reasoning steps to conclude the final answer.}
      \label{fig:workflow}
\end{figure*}



To study the effect of long-tail knowledge on LLM hallucinations in performing commonsense reasoning, we propose a large dataset for \underline{Co}mmonsense reasoning over \underline{Lo}ng-\underline{Ta}il entities (CoLoTa\footnote{\url{https://github.com/D3Mlab/CoLoTa}}) with 3,300 queries from both question answering and claim verification tasks. We find CoLoTa to be challenging even for the strongest LLMs. CoLoTa queries are formed by rewriting questions from the StrategyQA and claims from CREAK, replacing their popular entities with obscure ones from the Wikidata knowledge graph (KG). We observe that compared to the original queries, LLMs find commonsense reasoning on the long-tail queries far more challenging, 
resulting in notably higher hallucination rates as shown in Tables~\ref{tab:results-full} and~\ref{tab:results-small}. As shown in Figures \ref{fig:skillsqa} and \ref{fig:skillscv}, CoLoTa covers a diverse set of commonsense reasoning skills ranging from domain-independent skills like temporal reasoning and number comparison, to domain-specific skills such as reasoning about history, professions, sports, etc. Although the vulnerability of LLMs in memorizing facts about long-tail entities was observed in previous works~\citep{sun2023head}, the influence of long-tail knowledge on errors in commonsense reasoning and LLM hallucinations has been underexplored. We propose CoLoTa as a challenging and diverse benchmark for evaluating the commonsense reasoning capability of LLMs in long-tail settings.

The overall workflow of forming CoLoTa queries is shown with a worked example in Figure \ref{fig:workflow}. We select entity-based commonsense queries from StrategyQA and CREAK such that the factual knowledge required for answering them can be found in Wikidata. Next, we replace the popular entities targeted in the query with long-tail counterparts from Wikidata and annotate the relevant Wikidata facts, commonsense inference rules, and reasoning steps.

Since the support of knowledge required to answer CoLoTa queries is ensured to exist in Wikidata by design, CoLoTa can also serve as a novel benchmark for the task of answering natural language (NL) queries over KGs, known as Knowledge Graph Question Answering (KGQA). Early KGQA datasets consisted of simple questions that can be
answered using a single KG triple. Recent efforts in KGQA dataset
construction have focused on generating more complex queries
involving multi-hop reasoning; WebQuestions~\citep{berant2013semantic}, WebQuestionsSP~\citep{ yih2016value}, and GrailQA~\citep{gu2021beyond} are examples of such datasets. However, all these datasets focus on factoid questions, whereas more realistic queries as represented in CoLoTa often require both factual knowledge and commonsense knowledge and reasoning. 

In summary, CoLoTa not only (i) serves as a novel benchmark challenge dataset for LLM commonsense reasoning abilities and their robustness to hallucinations, but CoLoTa can also (ii) pave the way for future KGQA research in the LLM era to develop KGQA methods that incorporate factual and commonsense knowledge.


\section{Related Work}
\label{sec:related works}
In this section, we provide an overview of the commonsense reasoning and KGQA datasets and introduce exemplar datasets from each group. Exemplar queries from these datasets, their properties, and their comparison to CoLoTa are provided in Table~\ref{tab:dataset_comparison}.

\subsection{Commonsense Reasoning Datasets}

Humans naturally develop an understanding of how the world works, known as \emph{commonsense knowledge}, along with the ability to reason about it, called \emph{commonsense reasoning}~\citep{liu2004conceptnet}. These skills are fundamental to everyday thinking. For AI agents to engage meaningfully with the real world and interact with people, they must also acquire this form of knowledge~\citep {toroghi2024verifiable,baroni2017commai}. As a result, endowing AI agents commonsense knowledge and the ability to perform commonsense reasoning has been a core topic in AI research over the past decades~\citep{moore1982role, liu2004conceptnet, davis2015commonsense}. Due to this importance, several datasets are introduced to evaluate the commonsense reasoning capabilities of AI-based systems. In this section, we briefly summarize a number of widely used commonsense reasoning datasets by dividing them into two major categories: (i) Concept-based commonsense reasoning and (ii) entity-based commonsense reasoning datasets.

\begin{table*}[t!]
    \centering
    \small
    \caption{Examples of commonsense reasoning and KGQA datasets, their properties, and their comparison to CoLoTa. Answering CoLoTa queries requires incorporation of both factual knowledge and commonsense reasoning skills and can be used both to evaluate LLMs and KGQA methodologies.}
    \begin{tabular}{l l c c c}
        \toprule
        \textbf{Dataset Name} & \textbf{Exemplar Query} & \textbf{ Commonsense} & \textbf{Factual} & \textbf{Usable for KGQA} \\
        \midrule
        \multicolumn{5}{l}{\textbf{Commonsense Reasoning Datasets}} \\
        \midrule
        CommonsenseQA~\citep{commonsenseqa} & \textit{Where would I not want a fox? \textbf{A:} hen house} & \ding{51} & \ding{55} & \ding{55} \\
        \midrule
        PIQA~\citep{bisk2020piqa} & \textit{How do I find something I lost on the carpet?} & \ding{51} & \ding{55} & \ding{55}\\
         & \textit{\textbf{A:} Put a hair net on the end of your vacuum and turn it on.}&  & & \\
        \midrule
        SocialIQA~\citep{sap2019social} &  \textit{Tracy had accidentally pressed upon Austin} & \ding{51} & \ding{55} & \ding{55}\\
        & 
\textit{in the small elevator and it was awkward.}&  &  &  \\

 & \textit{why did Tracy do this? \textbf{A:} squeeze into the
elevator}&  &  &  \\

        \midrule
StrategyQA~\citep{strategyqa} &\textit{Could Tom Cruise explain mental auditing? \textbf{A:} Yes.} & \ding{51} & \ding{51}  & \ding{55}\\
\midrule
         
CREAK~\citep{creak} &\textit{All nuns act in holy ways. \textbf{A:} True.} & \ding{51} & \ding{51}  & \ding{55}\\
\midrule \midrule
        
        \multicolumn{5}{l}{\textbf{KGQA Datasets}} \\
        \midrule
        WebQuestions~\citep{webquestions}& \textit{What country is the Grand Bahama Island in? \textbf{A:}Bahamas} & \ding{55} & \ding{51} & \ding{51} \\
        \midrule
        LC-QuAD~\citep{lcquad} & \textit{What is the venue of Barack Obama’s marriage?} &  \ding{55} &  \ding{51} &  \ding{51} \\
         & \textit{\textbf{A:} Trinity United Church of Christ} &   & &  \\
         \midrule
        GrailQA~\citep{grailqa} & \textit{Which programming language did Andreas Rumpf develop?} & \ding{55}  & \ding{51}  & \ding{51}  \\
              & \textit{\textbf{A:} Nimrod} & &  &  \\
        \midrule
        \textbf{CoLoTa} & \textit{Would most people in New York be wearing jackets on} & \ding{51}& \ding{51} & \ding{51} \\
               & \textit{ Rajyavardhan Rathore’s birthday? \textbf{A:} Yes.} & & &  \\
        \bottomrule
    \end{tabular}

    \label{tab:dataset_comparison}
\end{table*}

\noindent
\textbf{ (i) Concept-based Datasets.}
Concept-based datasets evaluate commonsense reasoning skills by asking questions about general concepts or hypothetical scenarios. CommonsenseQA~\citep{commonsenseqa} is a notable benchmark in this group, consisting of multiple-choice commonsense questions. Queries of commonsenseQA are written by crowd workers who were provided a source concept and target concepts related to the source concept extracted from ConceptNet~\cite{concept-net}, a commonsense knowledge graph, and asked to write questions that discriminate between the target entities. ATOMIC~\citep{sap2019atomic} is another major work in this group, which is essentially an atlas of causal commonsense rules in the form of \textit{if-then} statements, that can be used to evaluate machine commonsense.

Another group of concept-based datasets evaluate specific commonsense reasoning skills. PIQA~\citep{bisk2020piqa} focuses on physical commonsense questions by expressing a goal and two physical activities as the options, one of which is plausible to achieve the goal. SocialIQA~\citep{sap2019social} introduces hypothetical scenarios about social interactions among fictional characters, and contains questions about the motivations and consequences of social behaviors, as well as emotional reactions.

\noindent\textbf{(ii) Entity-based Datasets.} Concept-based datasets focus on evaluating lifted commonsense reasoning capabilities. However, grounded commonsense reasoning about particular objects, entities, events, persons, etc. is also a critical capability, which is studied by entity-based commonsense reasoning datasets. As opposed to concept-based datasets whose queries only require commonsense knowledge, answering entity-based queries requires leveraging commonsense reasoning skills as well as properly incorporating factual information about the entities that are targeted in the query.

StrategyQA~\citep{strategyqa} is a major entity-based dataset consisting of a question with a True/False answer. In this dataset, the required commonsense reasoning steps to answer the query are implicit and should be inferred using a strategy. Each query is annotated with the reasoning steps and the evidence Wikipedia paragraphs that contain the factual information to answer each step. To form StrategyQA, crowd workers were asked to write strategy questions given an entity term, and adversarial models are continuously trained to collect increasingly difficult questions.
CREAK~\citep{creak} is another major entity-based dataset that contains claims about entities that are either correct or incorrect. Similar to StrategyQA, CREAK is also formed around Wikipedia and is written by crowd workers who are asked to write queries given an entity term. However, unlike StrategyQA, a model-in-the-loop approach is not used. 

The common underlying approach for building the above datasets is to have crowd workers write questions given one or more entities. Although this approach of priming crowd workers gives them more freedom to write diverse and challenging questions that require implicit reasoning, the chosen starting entities are often popular entities, about which abundant documents are present on the Internet and therefore in the training data of the LLMs. \emph{As we observe in the experiments, recent LLMs can easily provide factual and valid answers to these queries. In contrast for CoLoTa, we specifically focus on the long-tail knowledge that we observe to be much more challenging for modern LLMs (cf. ``Original'' vs. ``CoLoTa'' columns in Tables \ref{tab:results-full} and \ref{tab:results-small}).}

\subsection{KGQA Datasets}

Knowledge Graphs are widely adopted frameworks for representing relational information about entities, with their applications ranging from healthcare~\citep{rastogi2020personal} to recommendation~\citep{raza2024comprehensive, toroghi2023bayesian,tang2023logicrec}. Efficiently retrieving relevant information from KGs  has been a long-standing problem, for which specialized query languages like RQL~\cite{karvounarakis2002rql} and SPARQL~\citep{sparql} have been proposed. However, formulating queries in these languages requires expertise, making KGs less accessible to non-expert users~\citep{toroghi2024right}. In response, answering natural language queries by leveraging relevant KG information---the task known as KGQA ~\citep{yih2016value, zheng2017natural, berant2013semantic, toroghi2024bayesian}---has been explored to make KGs more accessible to a wider audience. 

Several datasets have been proposed to advance research in this field. Early KGQA datasets consisted of simple questions that could be answered using a single KG triple. Recent efforts in KGQA dataset construction have focused on generating more complex questions that involve multi-hop reasoning. For instance, WebQuestions~\cite{webquestions} is a KGQA dataset based on the Freebase~\cite{freebase} KG which is built by obtaining single-entity questions using the Google Suggest API and using crowd workers to pick the ones that are answerable using only the Freebase page for the question's subject entity. WebQuestionsSP~\cite{webquestions-sp} is an extension of WebQuestions with SPARQL annotations. LC-QuAD ~\cite{lcquad} is a dataset grounded in DBpedia~\cite{lehmann2015dbpedia}, built by: (1) generating formal queries using templates as well as selected entities and predicates, (2) rewriting them in natural language using templates written for each query template, and (3) asking crowd workers to rewrite the natural language template into a natural language question. 
GrailQA~\cite{grailqa} is grounded in Freebase, and is built using a 4-step process: (1) algorithmically generating logical forms from a KG, (2) converting them to pseudo-natural language questions, (3) using crowdsourcing to paraphrase the pseudo-natural language questions into natural language questions and (4) using crowdsourcing to cross-validate the questions.




The most common approach for building KGQA datasets is to first write logical template forms and then rewrite them as natural language questions. Since logical forms are meant to be executed over KGs, they are by nature ``factoid'', and thus the resulting questions will naturally be factoid as well, no matter how complex the logical form is. \emph{However, there is a rich space of questions requiring commonsense reasoning that to the best of our knowledge, has not yet been covered in any existing KGQA dataset. In CoLoTa, we aim to bridge this critical gap to encourage research on KGQA to answer queries involving both long-tail Wikidata entities and a range of commonsense reasoning skills (e.g., numeric, temporal, geographical).}

\subsection{Effect of Long-tail Knowledge on LLMs}
The fact that long-tail knowledge poses challenges to LLMs has been observed in a number of previous works. For instance, \citet{zhou2023devil} observed that LLMs that have shown advanced coding skills have a considerably inferior performance on tasks associated with infrequent labels compared to data samples of
frequent labels. Works such as \citet{sun2024head} and \citet{mallen2023not} construct new datasets from existing factual QA datasets and observe that LLMs are less capable of memorizing facts about long-tail knowledge. Similarly, \citet{kandpal2023large} observed correlational and causal relationships between the prevalence of documents in pre-training corpora and the performance of LLMs in answering factual questions about them.
These works have all focused on studying the influence of long-tail knowledge on the fact memorization ability of LLMs, but \emph{to the best of our knowledge, no existing work has studied the role of long-tail knowledge on the commonsense reasoning ability of LLMs and the related issue of hallucinations.
In contrast, CoLoTa is the first LLM Commonsense reasoning dataset that both (i) supports KGQA and also (ii) addresses the challenge of commonsense reasoning with long-tail knowledge that has not been studied in prior work.}


\begin{table*}[t!]
\caption{Exemplar queries from the existing entity-based commonsense reasoning datasets and their modified counterparts in CoLoTa: \tcr{red parts} indicate logic flaws or question clarity problems and \tcb{blue parts} show lack of a target entity or not targeting long-tail knowledge.}.
\small
\centering
\begin{tabular}{|l|l|}
\hline
\multicolumn{2}{|c|}{\textbf{Question Answering (Example 1)}} \\ 
\hline
 \textbf{Original Query} & \tcr{Did} \tcb{Francois Mitterrand} ever meet \tcb{Barak Obama} while they both held the position of President?\\ 
 \hline
 \textbf{CoLoTa Query}&  \tcr{Could} \tcb{Liau Hiok-hian} and \tcb{Virginia Raggi} have met while they both held the position of council member?\\
 \hline
 \textbf{Modifications} & \tcr{1-} The original question intends to ask the possibility of occurrence of a meeting rather than whether the meeting actually took\\ & place, but phrases it in a misleading manner.\\
 &\tcb{2-} The famous entities are replaced with long-tail counterparts.\\
 \hline
\multicolumn{2}{c}{}\\

\hline
\multicolumn{2}{|c|}{\textbf{Question Answering (Example 2)}} \\ 
\hline
 \textbf{Original Query} & \tcr{Could} \tcb{Tom Cruise} \tcr{explain} mental auditing?\\ 
 \hline
 \textbf{CoLoTa Query}&  \tcr{Is it likely} for \tcb{Julia Nickson-Soul} to \tcr{be familiar with} mental auditing ?\\
 \hline
 \textbf{Modifications} & \tcr{1-} The original question makes the implicit assumption that every person who practices a religion is able to explain its terminology \\&  which is not necessarily True. In the modified query, we replace it with \textit{being likely to be familiar} which is more accurate.\\
 &\tcb{2-} The famous entities are replaced with long-tail counterparts.\\
 \hline
\multicolumn{2}{c}{}\\

\hline
\multicolumn{2}{|c|}{\textbf{Claim Verification (Example 1)}} \\ 
\hline
 \textbf{Original Query} & \tcb{All nuns} \tcr{act in holy ways}.\\ 
 \hline
 \textbf{CoLoTa Query}& \tcb{ Léocadie Gascoin} \tcr{considered her job} to be holy.\\
 \hline
 \textbf{Modifications} & \tcr{1-} The original question is vague. 
 \textit{Acting in holy ways} is not a clear expression. In the modified query, we replaced it with a \\ & clear question.\\
 &\tcb{2-} The question does not target a particular entity. In the modification, we targeted it on a long-tail entity.\\
 \hline
\multicolumn{2}{c}{}\\

\hline
\multicolumn{2}{|c|}{\textbf{Claim Verification (Example 2)}} \\ 
\hline
 \textbf{Original Query} & \tcb{The civil engineer} designed \tcr{the new suspension bridge} in \tcr{the city}.\\ 
 \hline
 \textbf{CoLoTa Query}&  \tcb{Władysław Folkierski} had \tcr{likely learned} the knowledge required to design \tcr{suspension bridges}.\\
 \hline
  \textbf{Modifications} & \tcr{1-} The original claim is vague. It is not clear what civil engineer it refers to and targets a specific event for which no evidence is \\ &provided.\\
 &\tcb{2-} The original claim does not anchor a specific KG entity, but in the modification we fixed this.\\
 \hline
\end{tabular}
\label{tab:modifications}
\end{table*}

\section{CoLoTa Dataset}
CoLoTa is formed by rewriting queries from two entity-based commonsense reasoning datasets: StrategyQA~\citep{strategyqa} and CREAK~\citep{onoe2021creak}. 
Our rewriting aims are threefold: (1) to replace head entities with long-tail entities to study the impact on commonsense reasoning and hallucination; (2) to ensure that all long-tail entities are in Wikidata to support KGQA; and (3) to decompose and annotate the commonsense reasoning steps to keep track of the facts, reasoning skills, and rules of inference required to verify the query answer.

In the following, we first describe the dataset properties to achieve these aims followed by its detailed construction.

\begin{table}[t!]
\caption{Most frequent reasoning skills required for answering queries from CoLoTa's question answering task with exemplar questions for each skill.}
\small
\centering
\begin{tabular}{|c|c|} 
\hline
\textbf{Reasoning Skill} & \textbf{Exemplar Question}\\
\hline
Temporal Reasoning & \textit{Could Giovanni Battista Crespi have }\\ &   \textit{read L'apostolo popolare?} \\
\hline
Historical & \textit{Would it have been possible for }\\ &\textit{Maria de Ventadorn to speak to someone}\\  & \textit{100 miles away?}\\
\hline
Sports & \textit{Could Kim Dae-jung form}\\ &\textit{a polo team from his children?}\\
\hline
Number Comparison & \textit{Did Dariush Homayoun have longer}\\ & \textit{longevity than Bozorg Alavi?}\\
\hline
Geographical & \textit{Is Lowshan located south of Gubadly?}\\
\hline
\end{tabular}
\label{tab:skillstableqa}
\end{table}

\begin{table}[t!]
\caption{Most frequent reasoning skills required for answering queries from CoLoTa's claim verification task with exemplar claims for each skill.}
\small
\centering
\begin{tabular}{|c|c|} 
\hline
\textbf{Reasoning Skill} & \textbf{Exemplar Claims}\\
\hline
Professions & \textit{Martina Di Bari mostly uses her }\\ &   \textit{foot when doing her job.} \\
\hline
Historical & \textit{María Subíes Forcada's home}\\ &\textit{country was among the allied powers.}\\
\hline
Temporal Reasoning & \textit{Giovanni Battista Casti's works}\\ &\textit{may be influenced by}\\ &\textit{ Maria Grazia Lenisa's poems.}\\
\hline
Political & \textit{Hussein Ali Shido's party}\\ & \textit{supported anti-capitalism.}\\
\hline
Entity Comparison & \textit{Bogna Sobiech married an athlete who}\\ & \textit{does a different sport from her.}\\
\hline
\end{tabular}
\label{tab:skillstablecreak}
\end{table}

\subsection{Description}

CoLoTa has a total of 3,300 entity-based commonsense reasoning queries formed around entities from Wikidata. The dataset consists of two subsets for question answering and claim verification tasks each forming half of the dataset. Each entry of the dataset consists of the following constituents: (i) a query in the form of a question or a claim (ii) the unique Wikidata QIDs of anchor entities mentioned in the query, (iii) the relevant Wikidata sub-graph, containing the factual information required to answer the query (iv) an inference rule expressing the commonsense knowledge required to answer the query in the form of a natural language axiom, and (v) reasoning steps grounded on the KG triples to conclude the final answer. 

\noindent 
\textbf{Query}
A CoLoTa query is either a question expressed as an interrogative sentence, or a claim expressed as a declarative sentence. We use the term \textit{query} to refer to both questions and claims. 
Each query $q$ has a ground truth answer $a_q \in \{\textit{True}, \textit{False}\}$ which can be found using the KG facts and performing commonsense reasoning.

\noindent 
\textbf{Wikidata entities.} 
The query mentions a set of anchor entities selected from the Wikidata KG that are factually required for the answer.
For each anchor entity, we provide both its label and the unique Wikidata identifier (QID). These QIDs and labels can be used to retrieve the relevant facts required to answer the queries. 


\noindent 
\textbf{Inference rule.}
The inference rule is a logical statement in the
form of a natural language axiom that formally expresses the commonsense knowledge that is required to answer the question or verify the claim. In contrast to the existing datasets that consider inference using commonsense knowledge as an implicit part of the reasoning process, in order to enhance verifiability, we explicitly annotate the entry with the inference rules. These inference rules express the set of properties of entities targeted in the query and relations among them as conditions and premises 
for the reasoning process, such that if all premises are true, then the answer is true.
Formally, denoting the set of entities targeted in a query $q$ by $\mathcal{E}_{q} = \{e_{1,q}, ..., e_{|\mathcal{E}_{q}|,q}\}$, inference rule $I_q$ is a natural language representation of the First-Order Logic (FOL) expression
\begin{equation}
\left( \bigwedge\limits_{i=1}^{|\mathcal{P}|}\bigwedge\limits_{j=1}^{|\mathcal{E}|}P_i(e_j) \right) \land \left( \bigwedge\limits_{i=1}^{|\mathcal{F}|}\bigwedge\limits_{j=1}^{|\mathcal{E}|}F_i(e_j) 
\, \langle\mathit{op}_j^i\rangle \, e_j^i \right) \implies a_q,
\end{equation}
where $\mathcal{P} = \{ P_1, ..., P_{|P|} \}$ is the set of predicates, $\mathcal{F} = \{ F_1, ..., F_{|F|} \}$ is the set of functions, $\langle\mathit{op}_j^i\rangle \in \{ =, \neq, <, \leq, >, \geq \}$ is a (dis)equality or comparison operator if the function value is numeric, $e_j^i$ is the entity compared to the function evaluation, and $a_q$ is the answer. 

The assignment of each property and function to an entity is performed using one of the formats of \textit{"\{entity\} must \{have property or properties\} to \{conclusion\}"} or \textit{"If \{entity has the property\}, then \{the conclusion\}"} whichever results in a sentence that sounds more natural. For example, \textit{"If Belén Rodríguez is a girl from Latin America and is about to celebrate her 15th birthday, it would make sense for her to ask for a quinceañera."}


\begin{figure}
  \centering
  \includegraphics[width=0.88\linewidth]{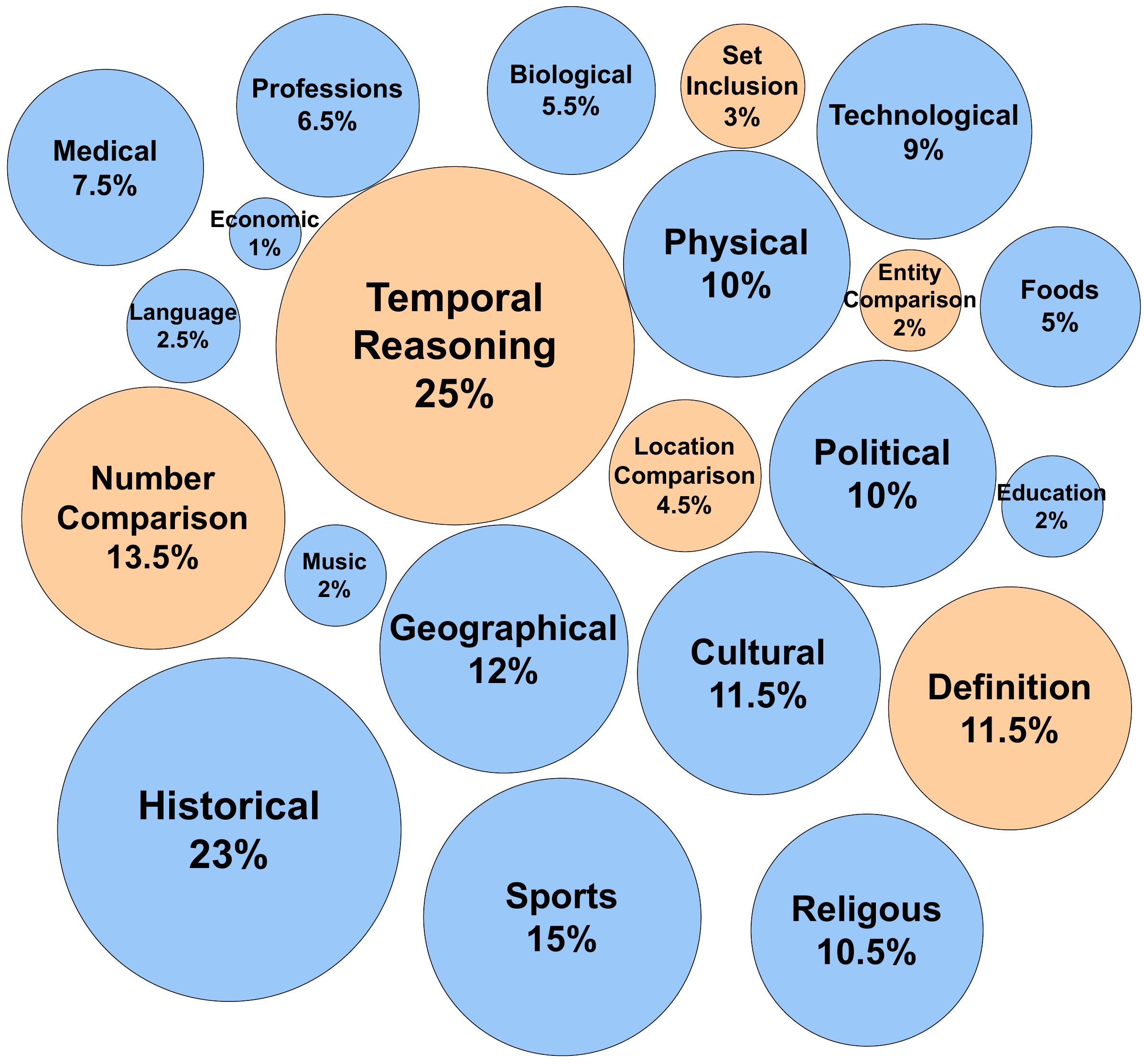}
    \caption{Distribution of reasoning skills in question answering task. Orange (Blue) circles show domain-independent (domain-dependent) reasoning skills.}
  \label{fig:skillsqa}
\end{figure}

\noindent 
\textbf{Relevant KG Sub-graph.}
Queries in CoLoTa are curated by ensuring that all factual information required for answering them is present in the Wikidata KG. For each query, we provide the set of ground-truth relevant triples in the relevant KG sub-graph of Wikidata that supply the required factual knowledge. 
Wikidata triples can include qualifiers in the format: \textit{relation-entity} pairs that provide additional context to the original KG triple. We show these hyper-relational triples as $\big((h, r, t), \{(k_i, v_i)\}_{i=1}^m\big)$ in which $h$, $r$, and $t$ denote the head, relation, and tail of the original triple, and $k_i$ and $v_i$ denote the relation and entity of the quantifier respectively. For example, for triple \textit{(Virginia Raggi, position held, mayor of Rome)}, an additional quantifier \textit{(replaces, Ignazio Marino)} is provided in Wikidata. We present the triple together with the quantifier as \textit{(Virginia Raggi, position held, mayor of Rome \{replaces, Ignazio Marino\})}.

\noindent 
\textbf{Reasoning steps.}
We decompose the commonsense inference rule into an ordered sequence of $n$ reasoning steps $S = (s^{(1)}, \dots, s^{(n)})$ for determining if each premise of the inference rule holds. There are two types of reasoning steps: (i) steps involving identifying the relevant facts among the facts retrieved from Wikidata, and (ii) steps in which logical reasoning about the extracted facts is performed.
Each reasoning step $s^{(i)}$ is written as a declarative sentence and may depend on the result of earlier steps $s^{(j)}, j < i$. Also, for each reasoning step $s^{(i)}$ that involves extraction of relevant facts, we provide the ground truth Wikidata triples $T^{(i)} = (T^{(i)}_1, \dots, T^{(i)}_m)$ that will help in evaluating the factuality of answers.

For each reasoning step $S^{(i)}$ that involves retrieving facts from Wikidata, we provide the ground truth evidence Wikidata triples $T^{(i)} = (T^{(i)}_1, \dots, T^{(i)}_m)$. Wikidata triples can contain qualifiers: key-value relation-entity pairs that provide additional context to the original KG triple. We represent these hyper-relational triples as: $\big((h, r, t), \{(k_i: v_i)\}_{i=1}^m\big)$

%
Unlike some existing KGQA datasets \cite{grailqa, webquestions-sp, lcquad}, we do not annotate CoLoTa with ground truth formal queries, since our questions require commonsense inferences and multiple steps of reasoning. Therefore, no single formal query can directly answer the question.

\subsection{Methodology}\label{method}

\noindent
\textbf{Query selection.}  In CoLoTa, we aim to form queries that are parallel to two existing entity-based commonsense reasoning datasets to assess the impact of entity popularity on the commonsense reasoning capability of LLMs. We choose two prominent entity-based commonsense reasoning datasets, StrategyQA~\citep{strategyqa} focusing on question answering and CREAK~\citep{onoe2021creak} that targets the claim verification task. 
Since CoLoTa is also intended to be used as a KGQA dataset, we select queries from these two datasets for which the necessary factual knowledge for their answers can be sourced in Wikidata. Having two sets of parallel queries, the original ones focusing on popular entities and the new ones forming CoLoTa targeting long-tail counterparts, we can study the influence of long-tail knowledge on commonsense reasoning capability (as demonstrated in the paired ``Original'' vs. ``CoLoTa''  columns of Tables~\ref{tab:results-full} and~\ref{tab:results-small}).

To select our queries, we first filter the queries to only include the ones whose anchor entities exist in Wikidata. Next, two annotators independently verify whether the required facts to answer each query exist in Wikidata. We keep queries that both annotators consider to satisfy this criterion. Since StrategyQA questions are annotated with strategy decompositions, we prioritized questions with greater numbers of strategy decompositions as they are more challenging to assess the LLM reasoning ability. For CREAK, claims with longer explanations involving more diverse reasoning skills were prioritized. We observed that claims of CREAK required an easier reasoning process compared to those of StrategyQA, which is also reflected in our experimental results in Section \ref{sec:experiments}. 


\begin{figure}
  \centering
  \includegraphics[width=0.79\linewidth]{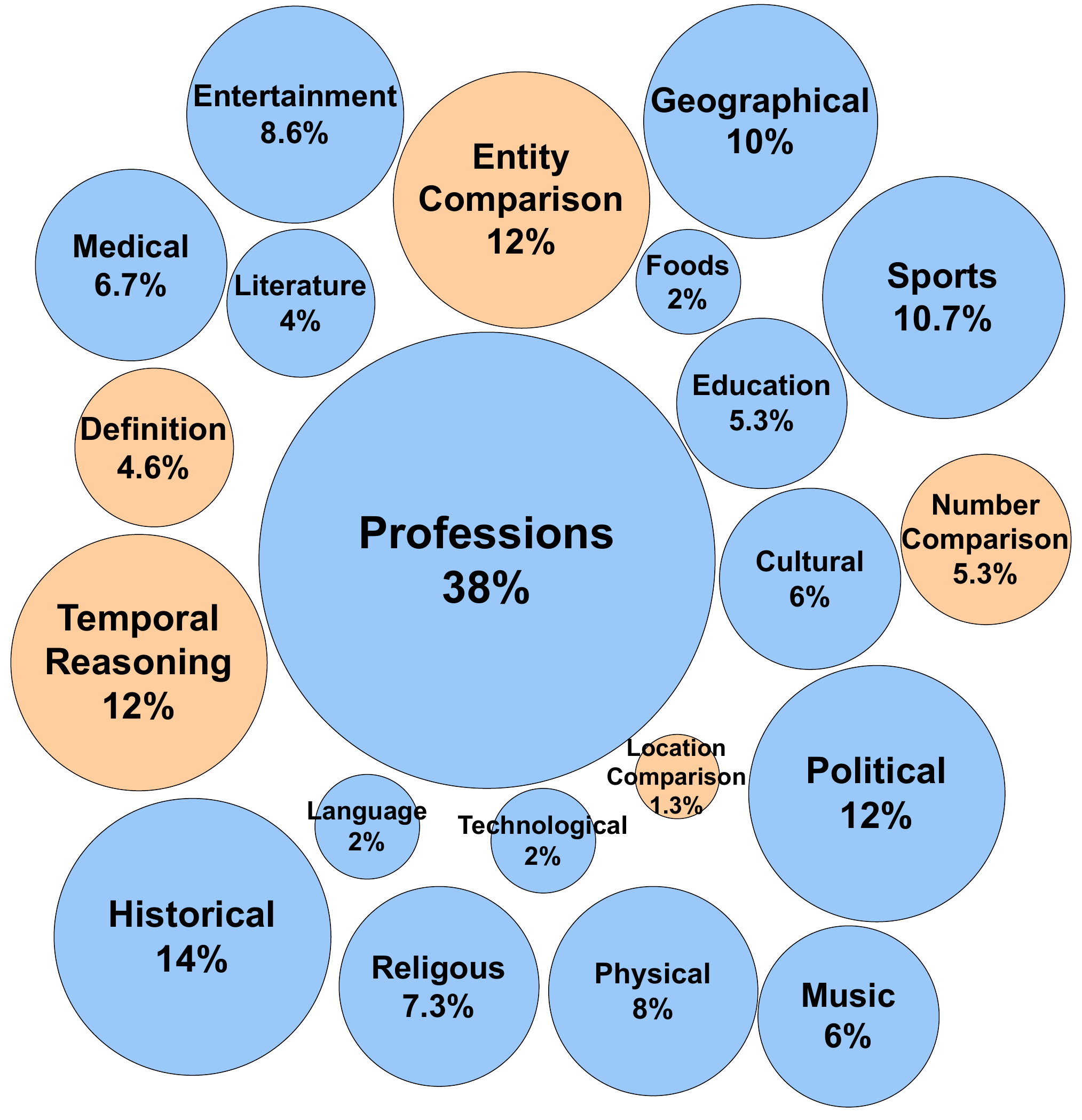}
\caption{Distribution of reasoning skills in claim verification task. Orange (Blue) circles show domain-independent (domain-dependent) reasoning skills.}
  \label{fig:skillscv}
\end{figure}

\noindent
\textbf{Entity substitution.} CoLoTa queries are designed to focus on obscure entities from long-tail knowledge. Hence, we replace the original popular entities with obscure counterparts. Specifically, we found similar counterparts of existing entities 
by manually generating SPARQL queries on Wikidata to retrieve candidate entities with similar properties to the original entity. Next, an entity with a much smaller number of Wikidata triples was randomly selected among the candidate entities. We modify queries that do not target any entity by introducing a long-tail KG entity. We also make some queries more challenging by adding commonsense reasoning indirection. For instance, in Figure \ref{fig:workflow}, instead of directly replacing \textit{``Tokyo''} with a long-tail city, we use another long-tail entity that was born in the substituted city as the anchor entity.

Examples of replacing popular entities with obscure counterparts are provided in Table \ref{tab:modifications}. 
Also, for queries with no particular target entity, we introduce long-tail entities and target the query on them.

\begin{figure}[!t]
    \centering

\includegraphics[width=0.77\linewidth]{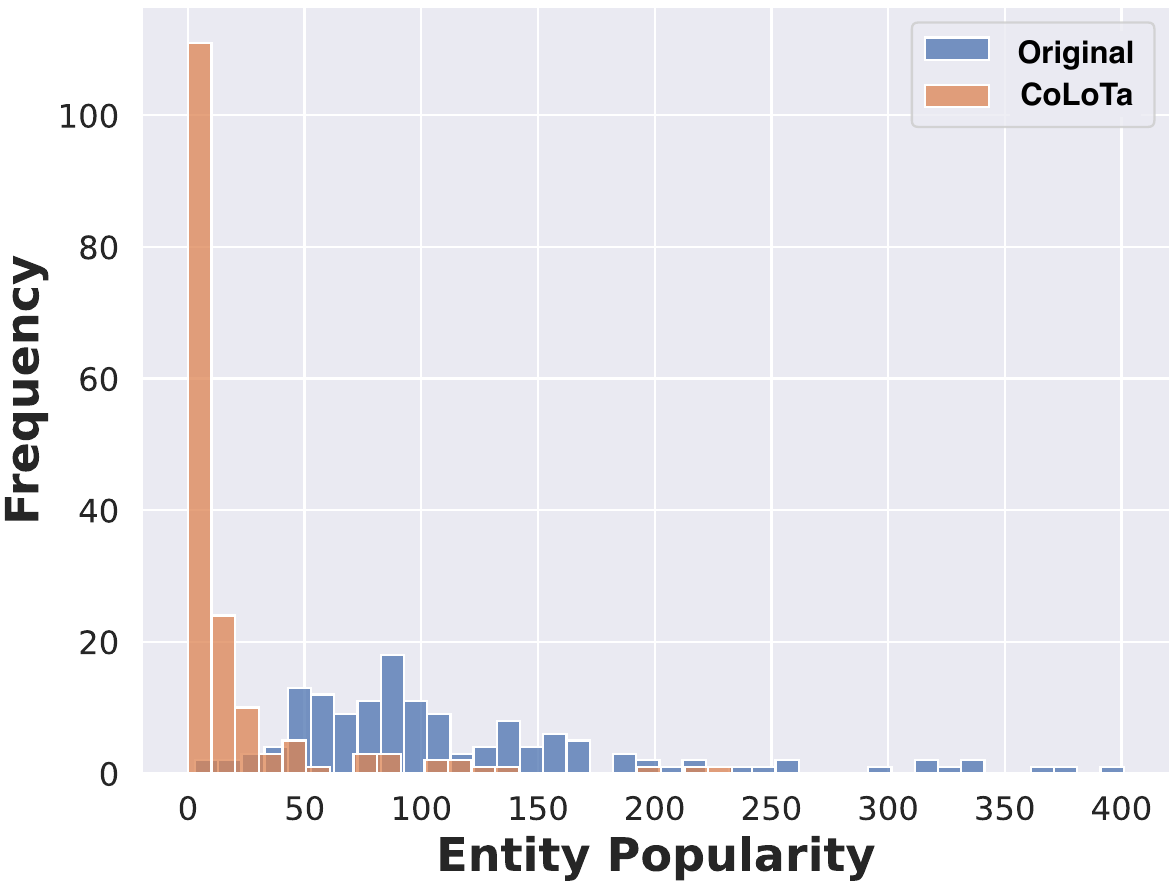} 
    \caption{Distribution of popularity of the entities targeted in claim
verification task of CoLoTa vs. the original queries, indicating CoLoTa's focus on long-tail entities.}
    \label{fig:popularity-cv}

\end{figure}

\begin{figure}[!t]
    \centering
   \includegraphics[width=0.77\linewidth]{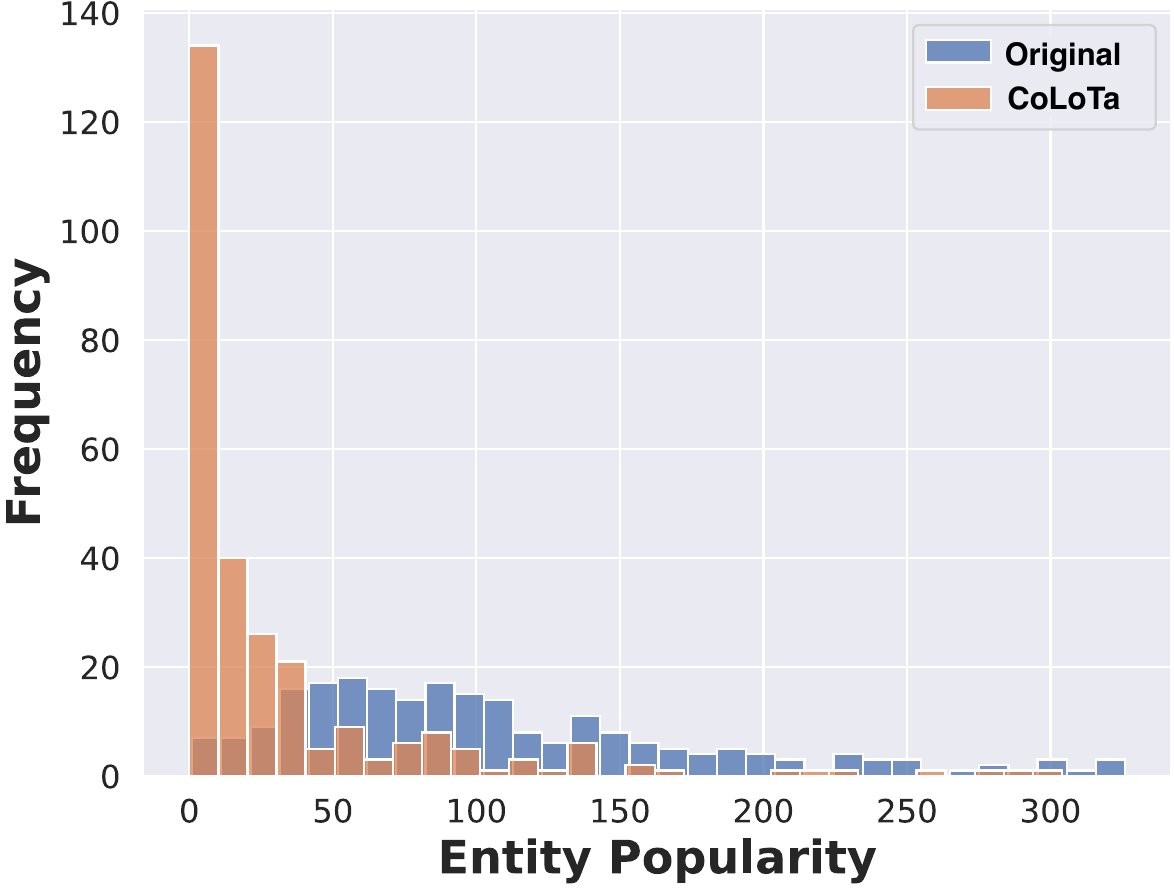} 
    \caption{Distribution of popularity of the entities targeted in question answering task of CoLoTa vs. the original queries, indicating CoLoTa's focus on long-tail entities.}
    \label{fig:popularity-qa}

\label{fig:popularities}
\end{figure}

To characterize the long-tail nature of our dataset, we plot the distribution of entity popularity for CoLoTa versus the original queries. 
Since our dataset is grounded in Wikidata, we use the number of Wikidata triples containing the entity as a measure for entity popularity. As shown in Figures \ref{fig:popularity-cv} and \ref{fig:popularity-qa}, the distributions of popularities of the entities targeted in CoLoTa queries are skewed toward smaller values, verifying the focus of CoLoTa on obscure entities that belong to the long-tail knowledge.

\noindent
\textbf{Query rewriting.} \citet{why-would-you-ask-it-that-way} found that questions in existing KGQA datasets are often worded unnaturally
, and showed that the performance of KGQA methods drop when evaluated on rewritten, more natural formulations of the same questions. 
We follow their scheme to 
rewrite our queries, which includes dimensions like grammar (e.g., poor word ordering, non-idiomatic) and form (e.g., quizlike, imperative phrasing). We also remove the incorrect implicit assumptions based on which some queries are written. Examples of the rewritten queries are provided in Table \ref{tab:modifications}.

\noindent
\textbf{Reasoning Skills.} 
To explore the diversity of reasoning skills required for solving the queries, following~\citep{strategyqa}, we annotate skills required for answering each query, which broadly fall under domain-specific (e.g., reasoning about geography, sports, etc.) and domain-independent (e.g., temporal reasoning, number comparison, etc.) categories. The distributions of the skills required for the question answering and claim verification tasks are shown in Figures \ref{fig:skillsqa} and \ref{fig:skillscv} respectively, in which the size of the circle for each skill is proportional to the percent of CoLoTa queries that require it. These figures indicate the broad coverage and diversity of the necessary skills across both tasks. Examples of queries from the five most frequent skills in each task are provided in Table \ref{tab:skillstableqa} and \ref{tab:skillstablecreak}.





\section{Experiments}
\label{sec:experiments}
\subsection{Baselines}
Since CoLoTa is intended to serve both as a dataset for evaluating entity-based commonsense reasoning capability of LLMs in long-tail settings, as well as a KGQA dataset, we use both LLMs and KGQA methodologies for evaluation. To study the impact of long-tail setting, we evaluate our studied methods on both original queries that mostly focus on popular queries and the long-tail counterparts from CoLoTa.

\noindent \textbf{LLM Baselines.} We evaluate five widely-used LLMs: GPT-3.5 Turbo, GPT-4o, OpenAI-o1, Gemini-1.5 Flash, and Llama-3.3-70B.
We evaluate these LLMs using both zero-shot~\citep{kojima2022large} and few-shot~\cite{cot} ($k=2$) Chain of Thought (CoT) prompting.

\noindent \textbf{KGQA Baselines.}
A majority of existing KGQA methods are only designed to answer factual queries by converting them to structured query languages such as SPARQL\citep{seaborne2008sparql}. 
These methods are generally not expected to be able to answer queries involving commonsense reasoning, such as CoLoTa's queries. However, recent KGQA methods leverage LLMs in their frameworks that may provide the required commonsense knowledge.

We use two strong LLM-based KGQA baselines to study whether the LLM can provide the required commonsense knowledge for the model. Our studied baselines include: (i) KB-Binder~\cite{kb-binder}, a semantic parsing method that leverages the in-context learning capability of an LLM with few-shot examples to convert natural language queries to SPARQL, and (ii) KGR~\cite{kgr} that uses an LLM to provide claims, corrects the LLM claims by retrofitting them on the KG, and then uses the LLM to suggest the final answer.

\subsection{Evaluation Metrics}
\noindent \textbf{Accuracy.} We use the accuracy of the final answers as the key indicator of each method's performance.

\noindent \textbf{Answer Rate.} Solely relying on accuracy to evaluate the question-answering performance of a model can be misleading. Especially in the long-tail setting of CoLoTa, where the model is likely to lack the factual knowledge to answer the query, a model may obtain higher accuracies by providing guessed or hallucinated answers that happen to be correct, compared to a model that refrains from answering those queries by providing responses like \textit{``I don't know''}. 

To distinguish incorrect answers from \textit{``I don't know''} answers, we also consider answer rate, the fraction of queries for which the method provides a \textit{True/False} answer.

\noindent
\textbf{FActScore.}~\citep{factscore} is a metric that measures the percentage of atomic facts in the LLM's response that are supported by a given knowledge source $C$, in our case Wikidata. Given a response $y$ consisting of a set of atomic facts $A_y$, the FActScore is defined as 
\begin{equation}
    f(y) = \frac{1}{|A_y|} \sum_{a \in A_y} \mathbb{I}(a \text{ is supported by } C).
\end{equation}

\noindent
\textbf{Reasoning score.} 
We define an analogous metric as FActScore, but for the validity of the reasoning logic. Given a LLM's response $y$ consisting of a sequence of reasoning steps $S = (s_1, \dots, s_n)$, step $s_i$ is valid $V(s_i) = 1$ if and only if $s_i$ can be logically deduced from all previous steps $s_j, j < i$. Thus, the reasoning process $S$ that forms the answer $y$ is valid ($V(S)=1$) if and only if all intermediate reasoning steps are correct, i.e., $V(S) = \bigwedge\limits_{i=1}^n V(s_i)$.

\begin{figure}[!t]
    \centering

\includegraphics[width=1\linewidth]{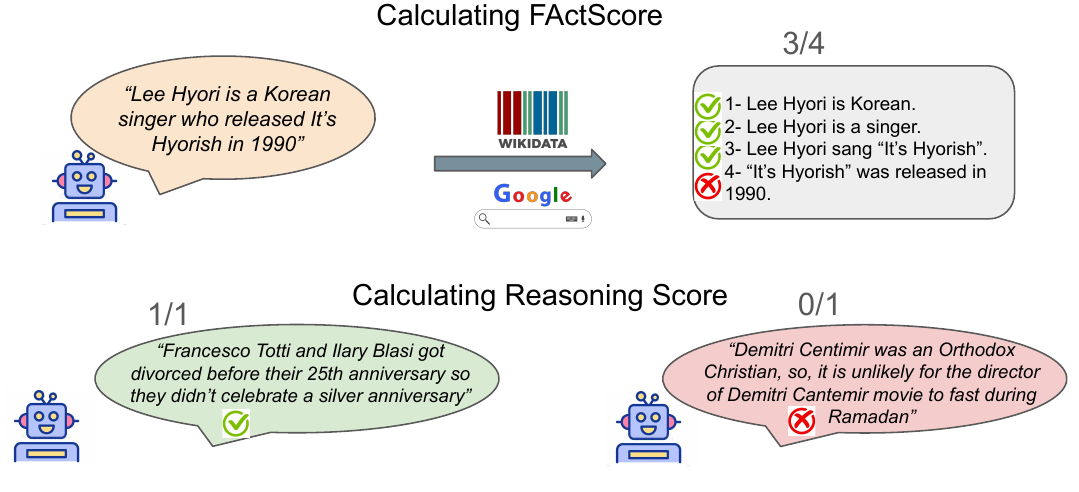} 
    \caption{Calculation of FActScore and Reasoning score for LLM responses through a worked example. }
    \label{fig:factscore}

\end{figure}

Calculation of the FActScore and reasoning score are also depicted through worked examples in Figure \ref{fig:factscore}.

\begin{table*}[h]
\small
\centering
\caption{Performance comparison of different models for question answering and claim verification tasks. Numbers in parenthesis indicate the amount of decrease in metric value for CoLoTa queries, compared to the original ones. It is notable that the accuracies obtained by all LLMs degrade when switching from Original to CoLoTa queries, whereas the reduction in answer rates is often much smaller, particularly for the best-performing LLM, OpenAI-o1. This effect indicates that although the LLMs are not able to provide correct answers to CoLoTa's queries, they still attempt to provide an answer, which leads to hallucinations and reasoning errors. }
\begin{tabular}{lcccccccc}
\toprule
\textbf{Model} 
& \multicolumn{4}{c}{\textbf{Question Answering}} 
& \multicolumn{4}{c}{\textbf{Claim Verification}} \\
\cmidrule(lr){2-5} \cmidrule(lr){6-9}
 & \multicolumn{2}{c}{\textbf{Accuracy}} & \multicolumn{2}{c}{\textbf{Answer Rate}} 
 & \multicolumn{2}{c}{\textbf{Accuracy}} & \multicolumn{2}{c}{\textbf{Answer Rate}} \\
\cmidrule(lr){2-3} \cmidrule(lr){4-5} \cmidrule(lr){6-7} \cmidrule(lr){8-9}
 & \textbf{Original} & \textbf{CoLoTa} 
 & \textbf{Original} & \textbf{CoLoTa} 
 & \textbf{Original} & \textbf{CoLoTa} 
 & \textbf{Original} & \textbf{CoLoTa} \\
\midrule
GPT-3.5-Turbo (Zero-shot CoT) & 0.76& 0.52 (0.24 $\downarrow$) & 1.00& 0.88 (0.12 $\downarrow$)& 0.93  & 0.58 (0.35 $\downarrow$)& 0.99& 0.86 (0.13 $\downarrow$) \\
GPT-3.5-Turbo (Few-shot CoT) & 0.79 & 0.52 (0.27 $\downarrow$) & 1.00& 0.82(0.18 $\downarrow$)& 0.93 & 0.59 
 (0.34 $\downarrow$)& 1.00 & 0.84 (0.16 $\downarrow$) \\
GPT-4o (Zero-shot CoT) & 0.83 & 0.64 (0.19 $\downarrow$)& 1.00& 0.97 (0.03 $\downarrow$)& 0.92 & 0.72 (0.20 $\downarrow$)& 1.00& 0.97 (0.03 $\downarrow$)\\GPT-4o (Few-shot CoT) & 0.85 & 0.64 (0.21 $\downarrow$)& 0.99 & 0.95 (0.04 $\downarrow$)& 0.92& 0.72 (0.20 $\downarrow$)& 1.00 & 0.96 (0.04 $\downarrow$)\\
OpenAI-o1 (Zero-shot CoT) & \textbf{0.88}& 0.67 (0.21 $\downarrow$)& 1.00 & 0.97 (0.03 $\downarrow$)& \textbf{0.93} & 0.71 (0.22 $\downarrow$)& 0.99 & 0.99 \\
OpenAI-o1 (Few-shot CoT) & 0.86& \textbf{0.68} (0.18 $\downarrow$)& 1.00 &0.94 (0.06 $\downarrow$)& \textbf{0.93}& \textbf{0.73} (0.20 $\downarrow$)& 1.00 & 0.97 (0.03 $\downarrow$)\\
Gemini-1.5 Flash (Zero-shot CoT) & 0.73 & 0.52 (0.21 $\downarrow$)&1.00 & 0.99 (0.01 $\downarrow$)& 0.92& 0.66 (0.26 $\downarrow$)& 1.00& 1.00 \\
Gemini-1.5 Flash (Few-shot CoT) & 0.77 & 0.57 (0.20 $\downarrow$)& 1.00 & 0.98 (0.02 $\downarrow$)& \textbf{\textbf{0.93}} & 0.70 (0.23 $\downarrow$)& 1.00 & 0.99 (0.01 $\downarrow$)\\
Llama-3.3-70B (Zero-shot CoT) & 0.65 &0.50 (0.15 $\downarrow$)& 0.86 & 0.70 (0.16 $\downarrow$)& 0.84 & 0.42 (0.42 $\downarrow$)& 0.97& 0.71 (0.26 $\downarrow$)\\
Llama-3.3-70B (Few-shot CoT) & 0.69& 0.46 (0.23 $\downarrow$)& 0.93& 0.66 (0.27 $\downarrow$)& 0.86& 0.47 (0.39 $\downarrow$)& 0.95 & 0.65 (0.30 $\downarrow$)\\
KGR & 0.39 &0.13 (0.26 $\downarrow$)& 0.51 & 0.17 (0.34 $\downarrow$)& 0.80 & 0.20 (0.60 $\downarrow$)& 0.85& 0.32 (0.53 $\downarrow$)\\
KB-Binder & 0.11& 0.08 (0.03 $\downarrow$)& 0.16& 0.14 (0.02 $\downarrow$)& 0.35& 0.14 (0.21 $\downarrow$)& 0.44 & 0.21 (0.23 $\downarrow$)\\
\bottomrule
\end{tabular}

\label{tab:results-full}
\end{table*}

\begin{table*}[h]
\small
\centering
\caption{Factuality and correctness of reasoning for OpenAI-o1 and GPT-3.5-Turbo LLMs. Numbers in
parentheses indicate the amount of decrease in metric value for CoLoTa queries, compared to the original ones. All metrics are smaller for CoLoTa queries, indicating their challenging nature that leads to a greater rate of factual hallucinations and reasoning errors.}
\begin{tabular}{lcccccccc}
\toprule
\textbf{Model} 
& \multicolumn{4}{c}{\textbf{Question Answering}} 
& \multicolumn{4}{c}{\textbf{Claim Verification}} \\
\cmidrule(lr){2-5} \cmidrule(lr){6-9}
 & \multicolumn{2}{c}{\textbf{FActScore}} & \multicolumn{2}{c}{\textbf{Reasoning}} 
 & \multicolumn{2}{c}{\textbf{FActScore}} & \multicolumn{2}{c}{\textbf{Reasoning}} \\
\cmidrule(lr){2-3} \cmidrule(lr){4-5} \cmidrule(lr){6-7} \cmidrule(lr){8-9}
 & \textbf{Original} & \textbf{CoLoTa} 
 & \textbf{Original} & \textbf{CoLoTa} 
 & \textbf{Original} & \textbf{CoLoTa} 
 & \textbf{Original} & \textbf{CoLoTa} \\
\midrule
GPT-3.5-Turbo (Zero-shot CoT) & 0.63& 0.54 (0.09 $\downarrow$)& 0.90& 0.89 (0.01 $\downarrow$)& 0.76  & 0.59 (0.17 $\downarrow$)& 0.93& 0.91 (0.02 $\downarrow$)\\
GPT-3.5-Turbo (Few-shot CoT) & 0.64 & 0.52 (0.12 $\downarrow$)& \textbf{0.92}& \textbf{0.90} (0.02 $\downarrow$)& 0.78 & 0.58 (0.20 $\downarrow$)& 0.93 & 0.92 (0.01 $\downarrow$)\\
OpenAI-o1 (Zero-shot CoT) & 0.95& \textbf{0.69} (0.26 $\downarrow$)& \textbf{1.00} & 0.83 (0.17 $\downarrow$)& 0.99 & 0.63 (0.26 $\downarrow$)& 1.00 & 0.88 (0.22 $\downarrow$)\\
OpenAI-o1 (Few-shot CoT) & \textbf{0.98}& 0.58 (0.40 $\downarrow$)& 0.95 &0.79 (0.16 $\downarrow$)& \textbf{1.00}& \textbf{0.70} (0.30 $\downarrow$)& \textbf{1.00} & 0.85 (0.15 $\downarrow$)\\

\bottomrule
\end{tabular}

\label{tab:results-small}
\end{table*}

\subsection{Results}

We now summarize key results from our experiments with CoLoTa:

\textbf{CoLoTa poses a significant challenge for the entity-based commonsense reasoning ability of state-of-the-art LLMs:} 
Accuracies and answer rate results for all baseline LLMs are reported in Table \ref{tab:results-full}. In brief, both metrics are lower for the CoLoTa queries across all methods, reflecting the fact that these methods find the long-tail queries in CoLoTa more challenging. 

The performance gap between the original queries and long-tail counterparts of CoLoTa is larger for the claim verification task. The drop in accuracies for the question answering task ranges between 0.15 and 0.27 among different LLMs, whereas this gap ranges from 0.20 to 0.42 for claim verification. In the most severe case, the accuracy of Llama-3.3-70B with zero-shot CoT drops by half (from 0.84 to 0.42).  
These results demonstrate that LLMs can adeptly perform commonsense reasoning about popular entities occurring in multiple training documents,
but struggle with long-tail ones. 

\textbf{Existing KGQA methods cannot perform commonsense reasoning to answer CoLoTa's queries:}
The LLM-based KGQA baselines, KGR~\citep{kgr}, and KB-Binder~\citep{kb-binder} perform abysmally on both the original and CoLoTa queries. The greatest accuracy that KB-Binder achieves is only 0.35 which is obtained on the original queries from the claim verification task. It also has a small answer rate because the execution of the SPARQL queries that it generates does not retrieve the required information from the KB, thus leading to a high rate of \textit{``I don't know''} answers. Compared to KB-Binder, KGR exhibits a stronger performance, reaching an accuracy of 0.80 on the original queries of the claim verification task. However, its performance on the CoLoTa queries drops to only 0.20. 
Even though these methods utilize LLMs that can provide commonsense knowledge,
we observe they cannot leverage the LLMs to perform commonsense reasoning. This makes CoLoTa a notable challenge dataset for improving this aspect of KGQA methods.

%

Among these baselines, OpenAI-o1, known for its strong reasoning capabilities~\citep{zhong2024evaluation}, often achieves greater accuracies. \emph{However, even OpenAI-o1 only achieves accuracies up to 0.68 and 0.73 on respective question answering and claim verification tasks of CoLoTa---much lower than the accuracies it achieved on the original queries.}

\textbf{LLMs still provide answers even when they lack knowledge of long-tail entities:}
A further critical observation is that although the accuracies of stronger LLMs such as OpenAI-o1 and GPT-4o drop on the CoLoTa queries, their answer rates remain very close to the original queries. This reflects the fact that although these state-of-the-art LLMs are unaware of the answers to CoLoTa's queries, they still try to make a guess, which can lead to hallucinations. 

\textbf{LLMs hallucinate on commonsense reasoning over long-tail entities:}
We study the factuality and validity of reasoning by calculating the FActScore and Reasoning score.
Since calculating these metrics requires manual verification of all intermediary steps for each response, we only calculate them for two methods: the strongest LLM (OpenAI-o1) and an LLM with moderate performance (GPT-3.5-Turbo), as reported in Table \ref{tab:results-small}. 
The factuality of answers and the correctness of reasoning are both smaller for the CoLoTa queries. OpenAI-o1 is a strong reasoner and shows near-ideal performance on the original queries, achieving a FActScore of 0.98 and a reasoning score of 0.95 on the question answering task and a FActScore and reasoning scores of 1.00 on the claim verification task. However, the factuality of its answers drops significantly on CoLoTa's queries, meaning that its correct answers are not necessarily grounded on correct facts, a clear indicator of increased hallucination rates. A similar drop is also observed in scores obtained by GPT-3.5-Turbo, as its FActScore drops up to 0.20.

\textbf{LLMs exhibit more reasoning errors in the long-tail setting:}
It is also important to note that the inferior performance of the studied LLMs on CoLoTa's long-tail queries is not only due to increased factual hallucinations—which may be intuitive as the LLMs are less familiar with the long-tail entities—but is also due to an increase in reasoning errors. This can be observed in the considerable drop in the reasoning score of OpenAI-o1 on both tasks of CoLoTa. Interestingly, the reasoning score of GPT-3.5-Turbo is more robust to the change of entity popularity than OpenAI-o1, but these metrics also drop slightly on CoLoTa queries. An example of such reasoning errors is provided in the incorrect example for calculating the reasoning score in Figure \ref{fig:factscore}, which is an output provided by GPT-3.5-Turbo. The LLM has made a logical error in its commonsense reasoning by claiming that if a person is an \textit{Orthodox Christian}, then the director of his biography movie is also likely not to \textit{fast during Ramadan}, which is logically flawed and incorrect.


\noindent \textbf{Results Summary.}  This notable increase in the rate of hallucinations as well as the severe drops in accuracies obtained by different state-of-the-art LLMs establishes CoLoTa as a challenging benchmark for studying LLM commonsense reasoning in long-tail settings and LLM hallucinations. Furthermore, the abysmal performance of strong KGQA methods on CoLoTa indicates its potential as a challenging dataset for future research on developing KGQA methodologies that are capable of commonsense reasoning.

\section{Conclusion}
We propose CoLoTa, a novel dataset for evaluating the entity-based commonsense reasoning capability of LLMs in long-tail settings. CoLoTa queries are formed by rewriting queries from existing commonsense reasoning datasets, replacing their popular entities with obscure counterparts.
We evaluated different LLMs using Chain of Thought prompting on CoLoTa, observing a high rate of hallucinations and reasoning errors even from LLMs that are known to be strong reasoners. 
Since the factual support of knowledge required for answering CoLoTa's queries is designed to exist in Wikidata, CoLoTa can also serve as the first KGQA dataset whose queries go beyond factoid retrieval and additionally require commonsense inferences. Experiments with two leading KGQA methods demonstrate their weak performance in answering CoLoTa's queries, indicating the insufficiency of existing KGQA methodologies to incorporate commonsense knowledge into their reasoning. Thus, CoLoTa serves as a challenging dataset for studying the commonsense reasoning ability of LLMs and their robustness to hallucinations, while also paving the way for future KGQA research in the LLM era. 
\begin{acks}
This work was supported by the Institute of Information \& Communications Technology Planning \& Evaluation (IITP) grant funded by the Korean Government (MSIT) (No. RS-2024-00457882, National AI Research Lab Project).
\end{acks}

\bibliographystyle{ACM-Reference-Format}
\bibliography{sample-base}

\end{document}